\documentclass{article}
\usepackage{ijcai18}
\usepackage{adjustbox}

\usepackage{times}
\usepackage{xcolor}
\usepackage{soul}
\usepackage[utf8]{inputenc}
\usepackage[small]{caption}
\usepackage[english]{babel}
\usepackage{graphicx}
\usepackage{mathtools}
\usepackage{graphicx}
\usepackage{subfig}
\usepackage{amsfonts}
\usepackage{amsfonts,amsthm,bm} 
\usepackage{hyperref}

\usepackage{algorithm}

\usepackage{algorithmic}
\usepackage{eurosym}

\usepackage{tikz}

\usetikzlibrary{decorations.pathreplacing}
\usepackage{sidecap}

\title{Machine learning modeling for time series problem:\\ Predicting flight ticket prices}

\author{
Jun Lu, 
\\ 
Computer Science, EPFL \\
jun.lu.locky@gmail.com,
}

\begin{document}

\maketitle

\begin{abstract}
Machine learning has been used in all kinds of fields. In this article, we introduce how machine learning can be applied into time series problem. Especially, we use the airline ticket prediction problem as our specific problem.

Airline companies use many different variables to determine the flight ticket prices: indicator whether the travel is during the holidays, the number of free seats in the plane etc. Some of the variables are observed, but some of them are hidden. 

Based on the data over a 103 day period, we trained our models, getting the best model - which is AdaBoost-Decision Tree Classification.  This algorithm has best performance over the observed 8 routes which has 61.35$\%$ better performance than the random purchase strategy, and relatively small variance over these routes.

And we also considered the situation that we cannot get too much historical datas for some routes (for example the route is new and does not have historical data) or we do not want to train historical data to predict to buy or wait quickly, in which problem, we used HMM Sequence Classification based AdaBoost-Decision Tree Classification to perform our prediction on 12 new routes. Finally, we got 31.71$\%$ better performance than the random purchase strategy.\footnote{A python implementation of this project is available online: \href{https://github.com/junlulocky/AirTicketPredicting}{https://github.com/junlulocky/AirTicketPredicting}}

\end{abstract}

\section{\bf{Introduction}}
For purchasing an airplane ticket, the traditional purchase strategy is to buy a ticket far in advance of the flight's departure date to avoid the risk that the price may increase rapidly before the departure date. However, this is usually not always true, airplane companies can decrease the prices if they want to increase the sales. Airline companies use many different variables to determine the flight ticket prices: indicator whether the travel is during the holidays, the number of free seats in the plane etc., or even in which month it is. Some of the variables are observed, but some of them are hidden. In this context, buyers are trying to find the right day to buy the ticket, and on the contrary, the airplane companies are trying to keep the overall revenue as high as possible. The goal of this article is to use machine learning techniques to model the behavior of flight ticket prices over the time.

Airline companies have the freedom to change the flight ticket prices at any moment. Travellers can save money if they choose to buy a ticket when its price is the lowest. The problem is how to determine when is the best time to buy flight ticket for the desired destination and period. In other word, when given the historical price and the current price of a flight for a specific departure date, our algorithms need to determine whether it is suitable to buy or wait. In order to build and evaluate the model, we use data that contain historical flight ticket prices for particular routes.

\section{\bf{Related work and our novelty}}
Some work has been done for determining optimal purchase timing for airline tickets. Our work is especially inspired by \cite{etzioni2003buy}. Described in the paper, it achieves 61.8$\%$ of optimal. This result is very close to our result. However, our project goes beyond their work in several ways: the observation period is over a 103 day period (instead of a 41 day period); we extracted 8 routes for the prediction (rather than 2 routes in the existing work). This is a more difficult problem because over a longer period, the airplane companies tend to vary the price algorithm behind the company and they may have different price strategies for different routes.

Moreover, our novelty is that we extended the problem into regression and classification problems by some \textbf{model constructions}. Finally, given the historical datas and current data of the ticket, our system can predict to buy or to wait. 

Another novelty is that, we also considered such a situation that some routes do not have any historical data, in which case we cannot perform the learning algorithms at all. This situation is very common, because there are always some new routes to be added by the airplane company or the company may conceal the historical data for some reasons. And also, this model can reduce computation time when we want to predict to buy or wait quickly because we do not need to train a large amount of data again. We call this problem as $\textbf{generalized problem}$. And on the contrary, for the previous problem, we call it as $\textbf{specific problem}$.

\begin{figure*}[h]
\centering
\includegraphics[width=16cm]{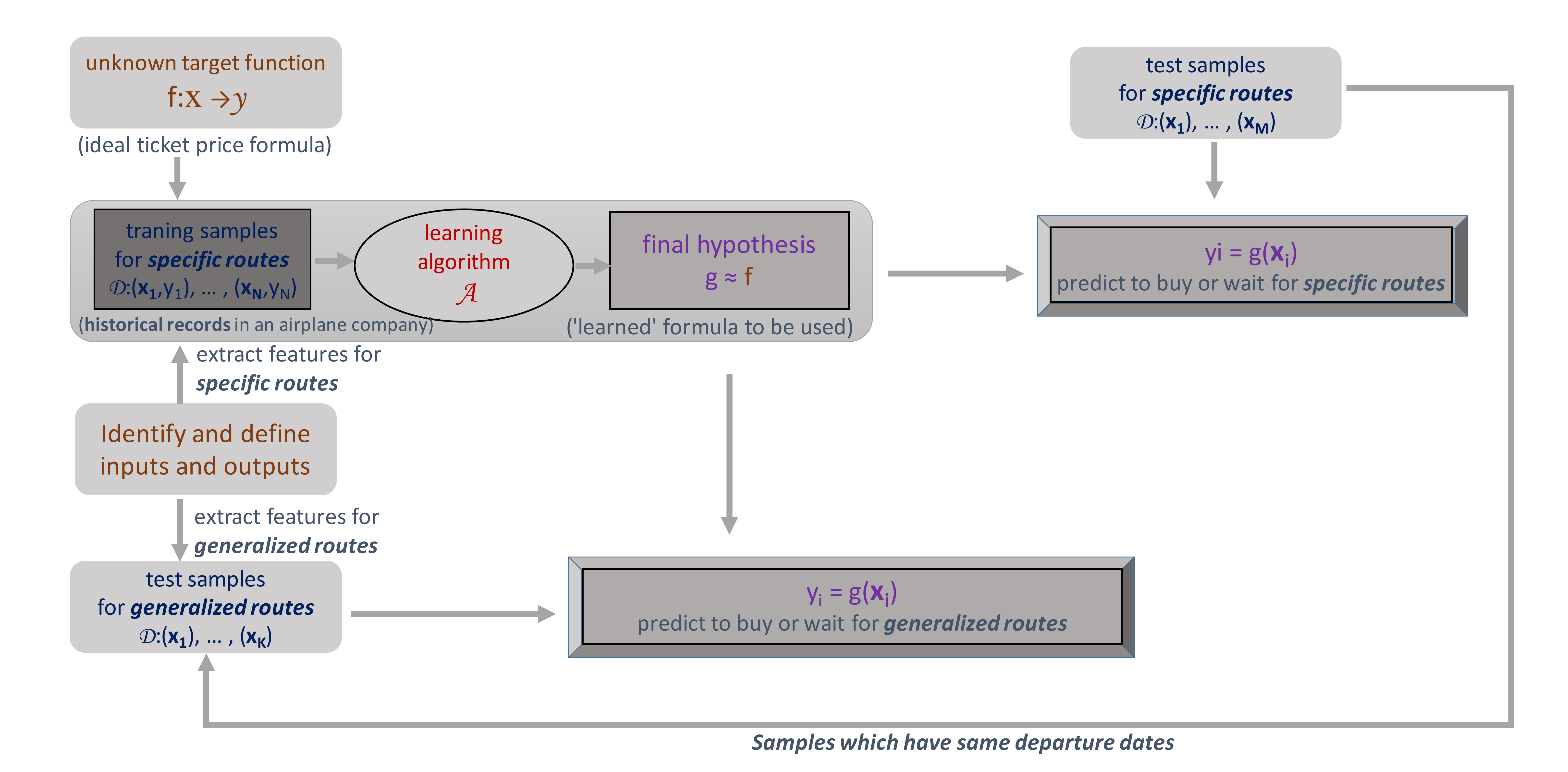}
\caption{Work flow for the specific problem and generalized problem.}
\label{genericworkflow}
\end{figure*}

In the generalized problem, we were not given any historical data for these routes. But we were given the formula learned from specific problem and corresponding data which has same departure data with specific problems (which will be described more clearly in the section of Data Description and Interpretation). So we need to extract input features based on the test samples of generalized routes and test samples of specific routes as shown in Fig. \ref{genericworkflow}. This model may have many benefits, especially when the historical data are not given or when we want to save time to quickly determine whether we should buy or wait for the new routes and not to spend too much time on model building for the new routes, and some other benefits such as decreasing data storage, interpretation and so on.

\section{\bf{Data Collection}}
The data for our analysis was collected as daily price quotes from a major airplane search web site between Nov. 9, 2015 and Feb. 20, 2016 (103 observation days). A web crawler was used to query for each route and departure date pair, and the crawling was done every day at 10:00 AM. 

For the purpose of our pilot study, we restricted the collecting data on non-stop, single-trip flights for 8 routes: 1 for Barcelona, Spain (BCN) to Budapest, Hungary (BUD); 2 for Budapest, Hungary to Barcelona, Spain; 3 for Brussels, Belgium (CRL) to Bucharest, Romania (OTP); 4 for Mulhouse, France (MLH) to Skopje, Macedonia (SKP); 5 for Malmo, Sweden (MMX) to Skopje, Macedonia; 6 for Bucharest, Romania to Brussels, Belgium; 7 for Skopje, Macedonia to Mulhouse, France; 8 for Skopje, Macedonia to Malmo, Sweden. And they are termed as R1 to R8. Overall, we collected 36, 575 observations (i.e. the queried price of each day for different departure dates and for the 8 different routes). In our observed airplane website, we did not find any tickets that were sold out in the specific queried days. 

For the generalized problem, we also collected another 12 routes (i.e. BGY$\rightarrow$OTP, BUD$\rightarrow$VKO, CRL$\rightarrow$OTP, CRL$\rightarrow$WAW, LTN$\rightarrow$OTP, LTN$\rightarrow$PRG, OTP$\rightarrow$BGY, OTP$\rightarrow$CRL, OTP$\rightarrow$LTN, OTP$\rightarrow$LTN, PRG$\rightarrow$LTN, VKO$\rightarrow$BUD, WAW$\rightarrow$CRL), which contains $14, 160$ observations and termed as R9 to R20. And you can notice that two routes have already been observed in the specific problem, which are CRL$\rightarrow$OTP and OTP$\rightarrow$CRL. We keep these two routes to see how the generalized model can influence the performance. We will compare the results of the specific problem and generalized problem for these two routes especially.

\subsection{\bf{Pricing Behavior in the Collected Data}}

\begin{figure}[!ht]
	\center
	\subfloat[Price change over time for flight R4, departing on Jan. 13 2016. This figure shows an example of price drops to the minimum a slightly before departure date and have more price fluctuation.]{
		\includegraphics[width=0.245\textwidth]{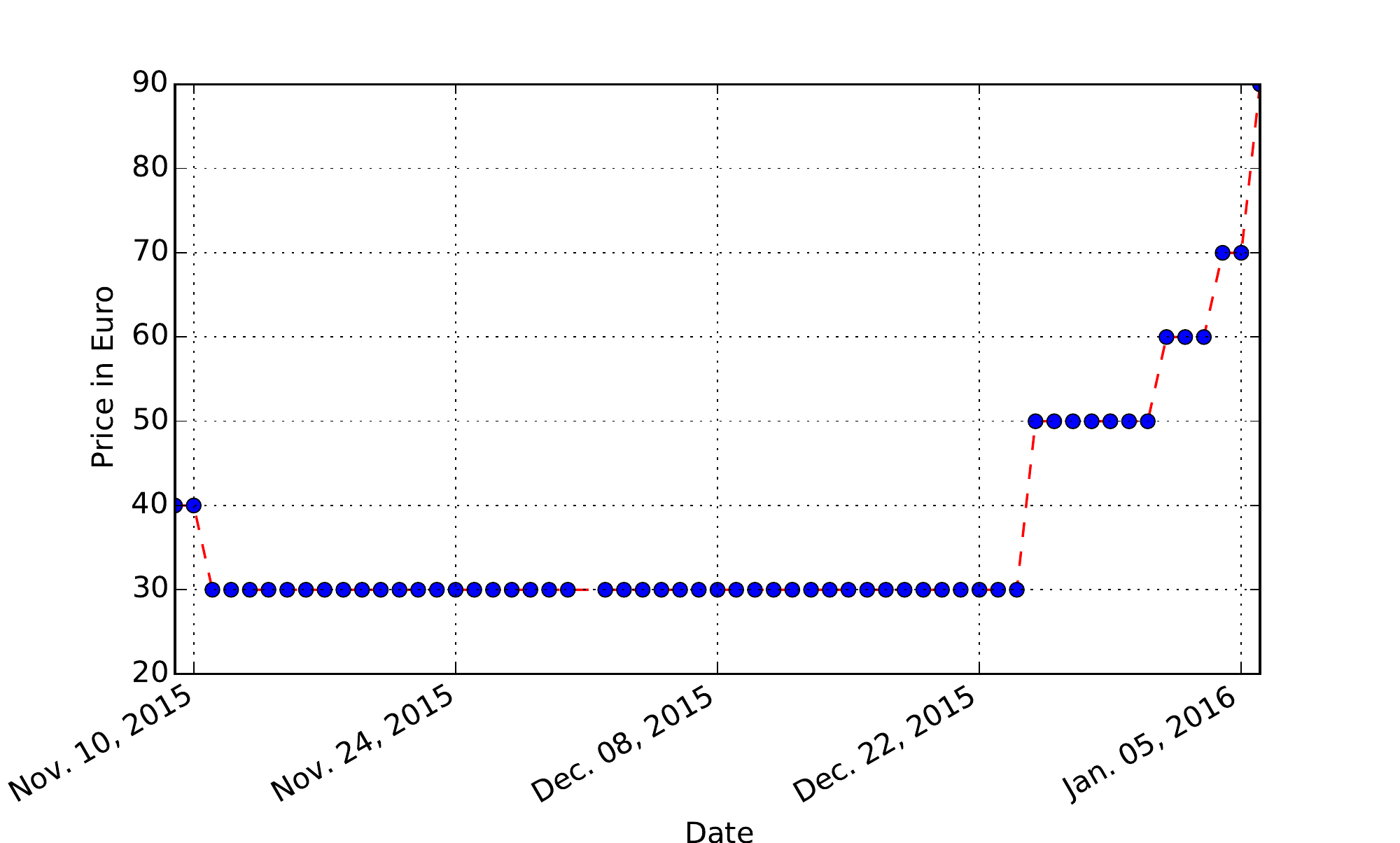}
		\label{price1}
	} 
	~
	\subfloat[Price change over time for flight R3, departing on Dec. 22 2015. This figure shows an example of price drops to the minimum a slightly before departure date which may benefit the consumers and the high prices dominate.]{%
		\includegraphics[width=0.245\textwidth]{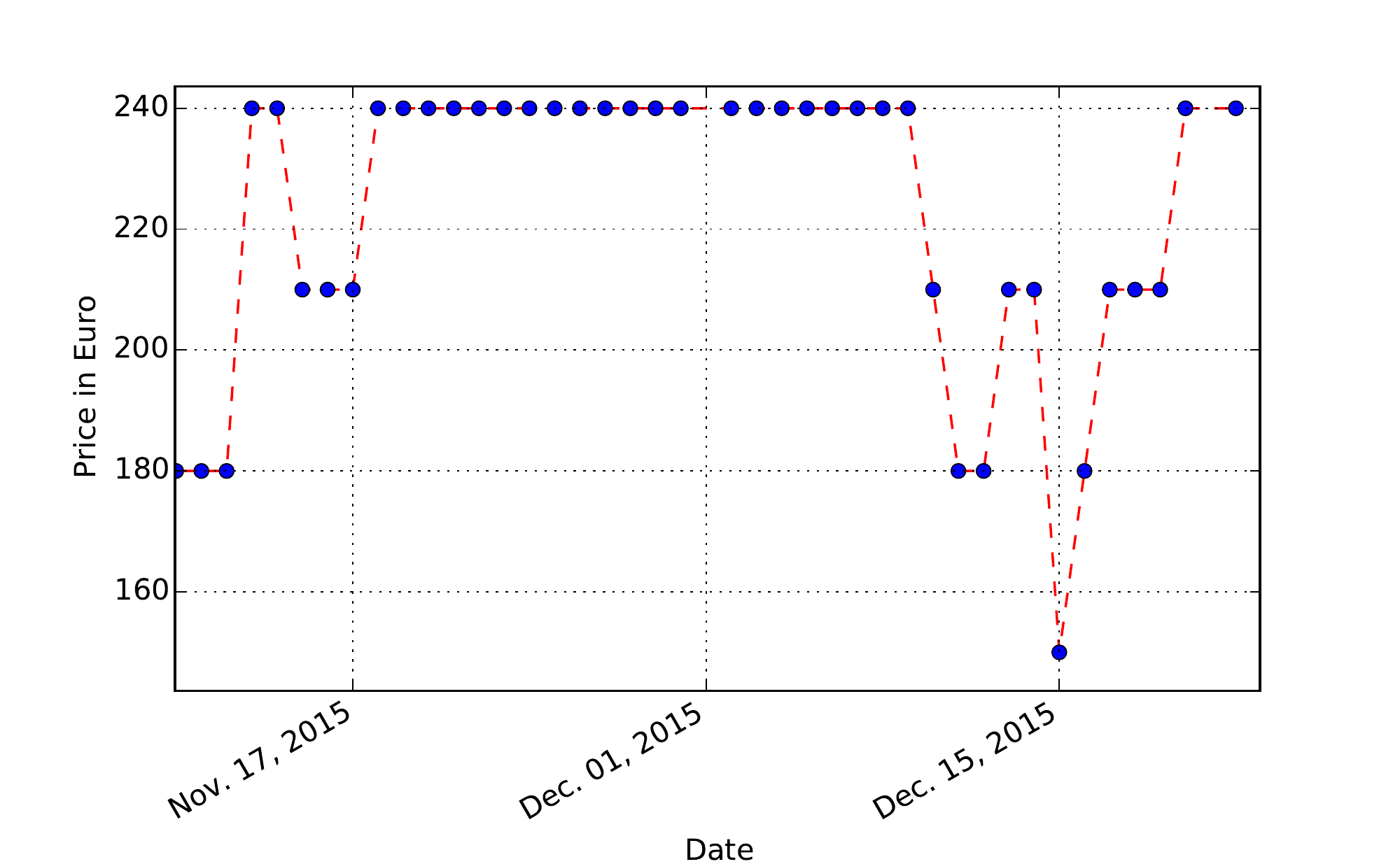}
		\label{price3}
	} 
	\caption{The price flow of two representative route.}
	\label{prices}
\end{figure}

We found that the ticket price for flights can vary significantly over time. Table \ref{priceChange} shows the minimum price, maximum price, and the maximum different in prices that can occur for flights of the 8 specific routes. In this table, although it's the maximum price and minimum price for all the departure dates for each route, we can have an overall glance at how we can achieve to decrease the ticket purchasing price. And Fig. \ref{price1} and \ref{price3} show how pricing strategies differ from flights.

\begin{table}[h!]
\centering
\begin{tabular}{|c|c|c|c|}
\hline
\textit{Route} & \textit{Min Price} & \textit{Max Price} & \textit{Max Price Change} \\ \hline
R1 & 29.99 \euro & 279.99 \euro & 250.0 \euro \\ \hline
R2 & 28.768 \euro & 335.968 \euro & 307.2 \euro \\ \hline
R3 & 9.99 \euro & 239.99 \euro & 230.0 \euro \\ \hline
R4 & 14.99 \euro & 259.99 \euro & 245.0 \euro \\ \hline
R5 & 15.48 \euro & 265.08 \euro & 249.6 \euro \\ \hline
R6 & 9.75 \euro & 269.75 \euro & 260.0 \euro \\ \hline
R7 & 16.182 \euro & 332.982 \euro & 316.8 \euro \\ \hline
R8 & 16.182 \euro & 332.982 \euro & 316.8 \euro \\ \hline
\end{tabular}
\caption{\bf{Minimum price, maximum price, and maximum change in ticket price for 8 specific routes. The decimals of the prices are from the currency change.}}
\label{priceChange}
\end{table}

%
%

\section{\bf{Machine Learning Approach}}
\subsection{\bf{Feature Extraction}}

The features extracted for training and testing are aggregated variables computed from the list of quotes observed on individual query days. For each query day, there are possibly eight airlines quoting flights for a specific origin-destination and departure date combination. For each query data, 5 features are computed, the $\emph{flight number}$ (encoded by dummy variables), the $\emph{minimum price so far}$, the $\emph{maximum price so far}$, the $\emph{query-to-departure}$ (number of days between the first query date (09.11.2015 in our case) and departure date), the $\emph{days-to-departure}$ (number of days between the query and departure date), and $\emph{current price}$. 

For the output for regression problem, we set it to be the minimum price for each departure date and each flight; and as for the output for classification problem, we set the data entry of which the price is the minimum price from the query date to the departure date to 1 (namely Class 1 - to buy), otherwise, we set it to be 0 (namely Class 2 - to wait).

\subsection{\bf{Data Description and Interpretation}}
Our data set consists of one set of input feature vectors, each one of them will be split into training dataset and testing dataset. In our case, we split the data that corresponds to flights with departure date in the interval between Nov. 9, 2015 and Jan. 15, 2016 as the training dataset; and the data that corresponds to flights with departure date in the interval betweeen Jan. 16, 2016 and Feb. 20, 2016 as the testing dataset; as for the generalized problem, we got the same time period as the test dataset, which is from Jan. 16, 2016 and Feb. 20, 2016.

Finally, the training dataset consists of $\bf{N_{tr}}$=16, 208 data samples of one output variable $\bf{y}$ and input variable $\bf{X}$. 
The testing dataset consists of $\bf{N_{te}}$=20, 367, for which the output is unknown, and where we forged our predictions.
The generalized problem testing dataset consists of $\bf{N_{ge}}$=14, 160, for which the output is unknown as well, and we need to predict.

\subsection{\bf{Model Construction}}
For regression, as the output is the minimum price for each departure date and each flight. Our regression method is to predict the expected minimum price for a given departure date and given flight with the input features. As a result, if the \textit{current price} is less than the \textit{expected minimum price}, we predict to buy; otherwise, we predict to wait. However, although it is very rare to happen, sometimes we may predict the \textit{expected minimum price} of every entry to be smaller than the \textit{current price}. Then, the last date should be seen as to buy. After we studied the data in depth, we were able to see that the last date always has a very high price. Thus we make the last buy date to be 7 days before departure date.

%
%
%
%
%
%

%
%
%

For classification, as our classification method is to predict to buy or to wait with the input features. As a result, if the prediction is 1, we buy the ticket, and we only buy the earliest. 

In our approach, we used 7 machine learning models to compare the results, namely Least Squares for Regression \cite{harvey1966least}, Logistic Regression for Classification \cite{dayton1992logistic}, 3 layer Neural Networks, Decision Tree \cite{smith2004decision}, AdaBoost-Decision Tree \cite{freund1999alternating}, Random Forest \cite{liaw2002classification}, K Nearest Neighbors \cite{peterson2009k}, Uniform Blending \cite{lin2016mlt} and Q Learning \cite{watkins1992q,etzioni2003buy}. For both the regression and classification problems, we tuned the hyperparameter via 5-fold cross validation (CV).

\subsection{\bf{Performance Benchmarks}}
The naive purchase algorithm, called the $\emph{Random}$ $\emph{Purchase}$, is to purchase a ticket randomly before the departure date. To be more concrete, for every departure date, for example departure date B, then from the first historical data for this departure date (namely date A), we pick several tickets randomly in this interval to simulate the clients buying ticket. And the average price would be computed as the $\emph{Random}$ $\emph{Purchase}$ $\emph{Price}$. Another purchase strategy benchmark introduced in \cite{groves2013optimal} is called $\textit{earliest purchase}$, in which it splits the time interval into many day periods and it purchases one ticket in each day period. These two performance benchmarks are similar. However, we thought that the $\textit{Random Purchase}$ strategy conforms more closely to reality because in reality some clients may choose the same time period to buy rather than buy tickets in separated time interval. 

The lowest achievable cost is called the $\emph{Optimal}$ $\emph{Price}$ and it is the lowest price between the first query date and the departure date.

\section{\bf{Generalized Model}}
\subsection{\bf{Uniform Blending}}
The simplest way to do the generalized predicting problem is using uniform blending. As long as we trained the specific problem, we could get 8 models (or learners) for 8 flight numbers separately. After that we can use the 8 models to predict for our new route, then we let these 8 models vote to buy or wait for every ticket. 

\subsection{\bf{HMM Sequence Classification}}
Apart from uniform blending, we can also allocate every data entry a "flight number" from the 8 routes, rather than average the 8 models. We then used sequence classification to allocate the "flight number" to every data entry. 
%
%
%
%

In our problem, given a new observation sequence and a set of models, we want to explore which model explains the sequence best, or in other terms which model gives the highest likelihood to the data so that to extract corresponding features for the entry. And then, we can plugin the features from this chosen route to the new sequence to predict.

\subsubsection{HMM sequence classification}
Also called Maximum Likelihood classification. In practice, it is very often assumed that all the
model priors are equal (i.e. that the new route to be predicted have equal probabilities of having same pattern in the observed 8 specific routes). Hence, this task consists mostly in performing the Maximum Likelihood
classification of feature sequences. For that purpose, we must
have of a set of HMMs \cite{eddy1996hidden} that model the feature sequences. These models can be considered as
\textbf{``stochastic templates''}. Then, we associate a new sequence to the most
likely generative model. This part is called the {\em decoding} of the
 feature sequences.

\subsubsection{\bf{Use HMM to solve our problem}}
\begin{figure}[h]
\centering
\includegraphics[width=8cm]{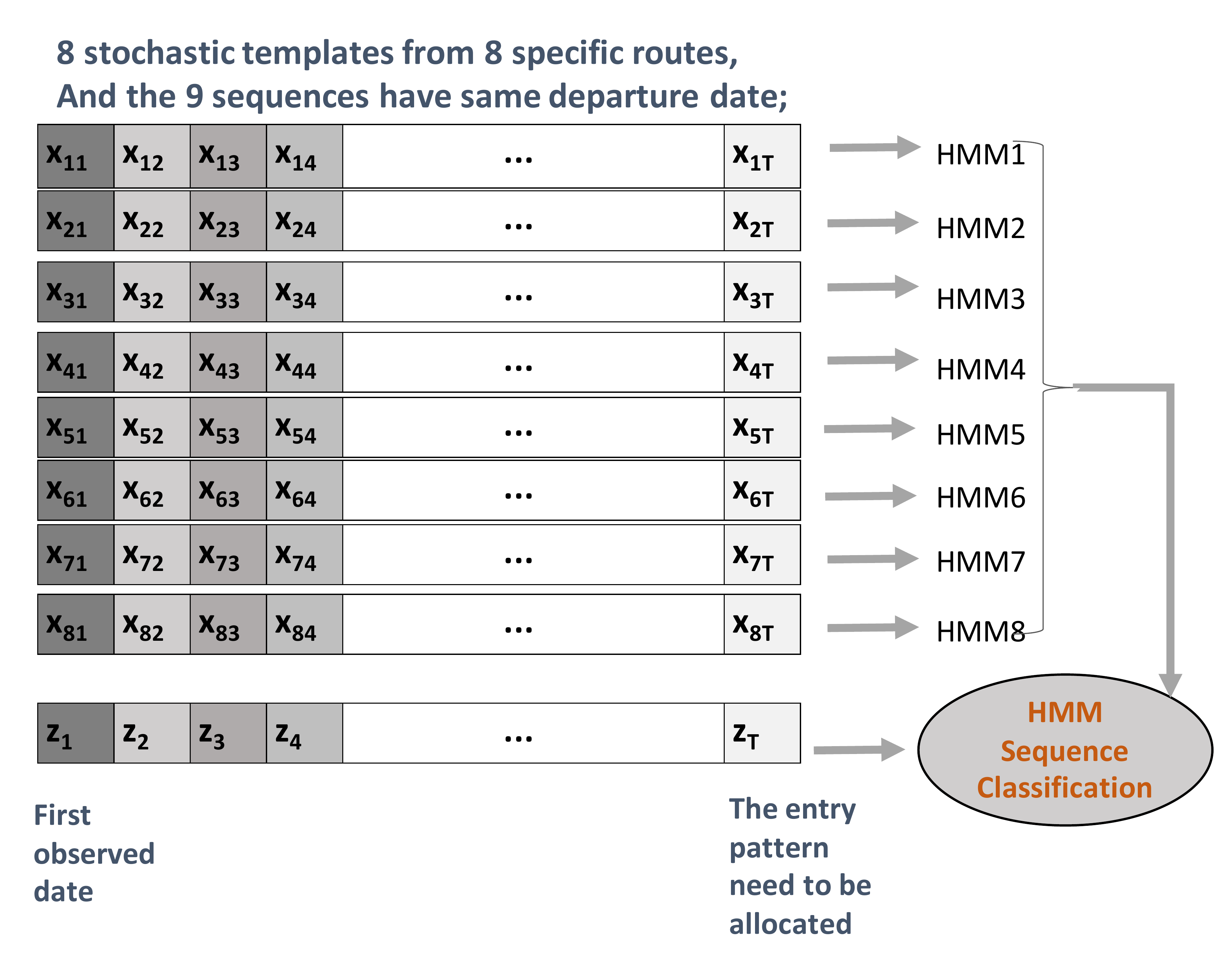}
\caption{How to use 8 stochastic templates in HMM sequence classification.}
\label{hmmsequence}
\end{figure}

We defined an \textbf{equivalence sequence} over different routes. Our equivalence sequence is the set of states with the same departure date, the same days before takeoff (i.e. current date or current entry to predict), and the same first observed date. Fig \ref{hmmsequence} shows how these 8 stochastic templates can be used to predict and intuitive meaning of \textbf{equivalence sequence}. Our goal is to allocate one pattern to each entry. 

In our case, as you know, we already have 8 patterns from the 8 specific routes. When we have to find out which pattern should be classified to the entries for a new route, we do a HMM Sequence Classification on the new sequence. In other word, we trained the 8 referenced sequence by HMM to get each HMM model parameters $\Theta_i$ (i=1, 2, ..., 8), i.e. getting the transition matrix and emitting probability of each model. Then we compared it to the new sequence to get the {\em likelihood of the new sequence with respect to each HMM model}. Finally, allocate the pattern to the entry which makes it have the largest probability. This model is widely used in Automatic Speech Processing. To be more concrete, if an data entry is allocated to the second pattern of the 8 patterns. Then the dummy variable of flight number allocated to this entry should be: $F$ =  [0,1,0,0,0,0,0,0]$^\top$. 



\subsection{\bf{Performance Metric}}
As long as we get the $\emph{Randome}$ $\emph{Purchase}$ $\emph{Price}$, $\emph{Optimal}$ $\emph{Price}$, and the $\emph{Predicted}$ $\emph{Price}$, we can use the following performance metric to evaluate our results:
\begin{equation} \label{perfor1}
\begin{split}
\emph{Performance} = \frac{\emph{Random Purchase Price} - \emph{Predicted Price}}{\emph{Random Purchase Price}}\% 
\end{split}
\end{equation}

\begin{equation} \label{perfor2}
\begin{split}
\emph{Optimal Perfor} = \frac{\emph{Random Purchase Price} - \emph{Optimal Price}}{\emph{Random Purchase Price}}\% 
\end{split}
\end{equation}

\begin{equation} \label{perfor3}
\begin{split}
\emph{Normalized Performance} = \frac{\emph{Performance}}{\emph{Optimal Performance}}\% 
\end{split}
\end{equation}

Having these metrics in mind, we used the Normalized Performance to evaluate our results, because it normalizes every route and it ranges from 0$\%$ to $100\%$, in which case, the higher the better, so that it gives more intuition about how well or bad is the result performance.

\section{Experiments}
\subsection{\bf{Regression Results}}
\begin{table*}[h]
  \centering
\begin{tabular}{| c | c | c | c | c | c | c | c| c| c|} 
\hline Performance($\%$)& \multicolumn{9}{|c|}{Model}  \\
\cline{1-10} Routes $\backslash$ Method & Optimal  & \begin{tabular}[c]{@{}l@{}}Random\\ Purch.\end{tabular} &  \begin{tabular}[c]{@{}l@{}}Linear\\ Regression\end{tabular} & NN & \begin{tabular}[c]{@{}l@{}}Decision\\ Tree\end{tabular} & KNN & AdaBoost & \begin{tabular}[c]{@{}l@{}}Random\\ Forest\end{tabular} & \begin{tabular}[c]{@{}l@{}}Uniform\\ Blending\end{tabular}\\ 
\hline 
\hline 
R1 & 100.0 & 0.00 & -42.17 & 67.03 & 42.31 & 38.19 & 48.49 & 54.67 &  44.37\\ 
\hline 
R2 & 100.0 & 0.00 & 1.91 & 57.12 & 72.66 & 91.96 & 71.59 & 96.78 & 80.17\\ 
\hline 
R3 & 100.0 & 0.00 & 10.34 & 18.60 & 30.40 & 37.48 & 41.01 & 56.35 & 25.68\\ 
\hline 
R4 & 100.0 & 0.00 & -21.02 & -10.74 & 22.37 & 36.07 & 31.50 & 77.17 & 29.22\\ 
\hline 
R5 & 100.0 & 0.00 & -118.53 & 53.25 & 16.46 & 67.91 & 53.80 & -0.69 & 65.70\\ 
\hline 
R6 & 100.0 & 0.00 & -19.47 & 57.30 & 63.85 & 69.09 & 64.11 & 67.51 & 63.85\\ 
\hline 
R7 & 100.0 & 0.00 & 47.72 & 35.80 & 50.11 & 52.10 & 51.30 & 45.34 & 45.34\\ 
\hline 
R8 & 100.0 & 0.00 & -96.14 & 64.63 & 43.33 & 30.07 & 43.33 & 46.34 & 49.76\\ 
\hline 
\hline
Mean Perf. & 100.0 & 0.00 & -29.67  & 42.87 & 42.68 & 52.86 & 50.64 & \bf{55.43} & 50.51\\ 
\hline
\hline
Variance & 0.00 & 0.00 & 2654.78  & 636.21 & \bf{37.04} & 409.11 & \bf{143.49} & 707.99 & 302.90\\ 
\hline
\end{tabular}
 \caption{\bf{Regression Normalized Performance Comparison of 8 routes.}}
 \label{regressionResult}
\end{table*}

Table \ref{regressionResult} shows the results of regression methods. Random Forest Regression gets the best performance in regression method. However, it's variance is not small enough, which means it is sensitive to different routes. In this case, although for some routes, it gets good performance, for other routes, it gets bad performance. From the aspect of the clients, it is not fair for the some clients to buy tickets for which the system may predict badly. The preferred method in regression is AdaBoost-Decision Tree Regression method, which has smallest variance and a relative high performance.

\subsection{\bf{Classification Results}}
\subsubsection{\bf{Solving Imbalanced Data Set}}
Concerning a classification problem, an imbalanced data set leads to biased decisions towards the majority class and therefore an increase in the generalization error. As referred in the section of Data Description and Interpretation of the classification problem, the number of samples per class in our problem is not equally distributed, only few entries are to buy, most of the entries should be to wait. To address this problem, we can consider three approaches, namely $\textbf{Random Under Sampling}$ (Randomly select a subset the majority classes' data points), $\textbf{Random Over Sampling}$ (Randomly add redundancy to the data set by duplicating data points of the minority classes) and $\textbf{Algorithmic Over Sampling}$ (Add redundancy to the data set by simulating the distribution of the data)

Having these methods in mind, firstly, the \emph{Random Under Sampling} would not be useful in our problem, because in our problem, the data set is very unbalanced, i.e. the buy entries is very sparse. If we use \emph{Random Under Sampling} \cite{japkowicz2002class,japkowicz2000class,wasikowski2010combating,guo2008class}. Secondly, if we use \emph{Algorithmic Over Sampling}, it will add many noises into the data, because we do not know the hidden relationship between the features and the output. As a result, we preferred the second method, which is \emph{Random Over Sampling}.

\subsubsection{\bf{Identification of Outliers}}
\begin{figure}[h]
\center
\includegraphics[width=6.8cm]{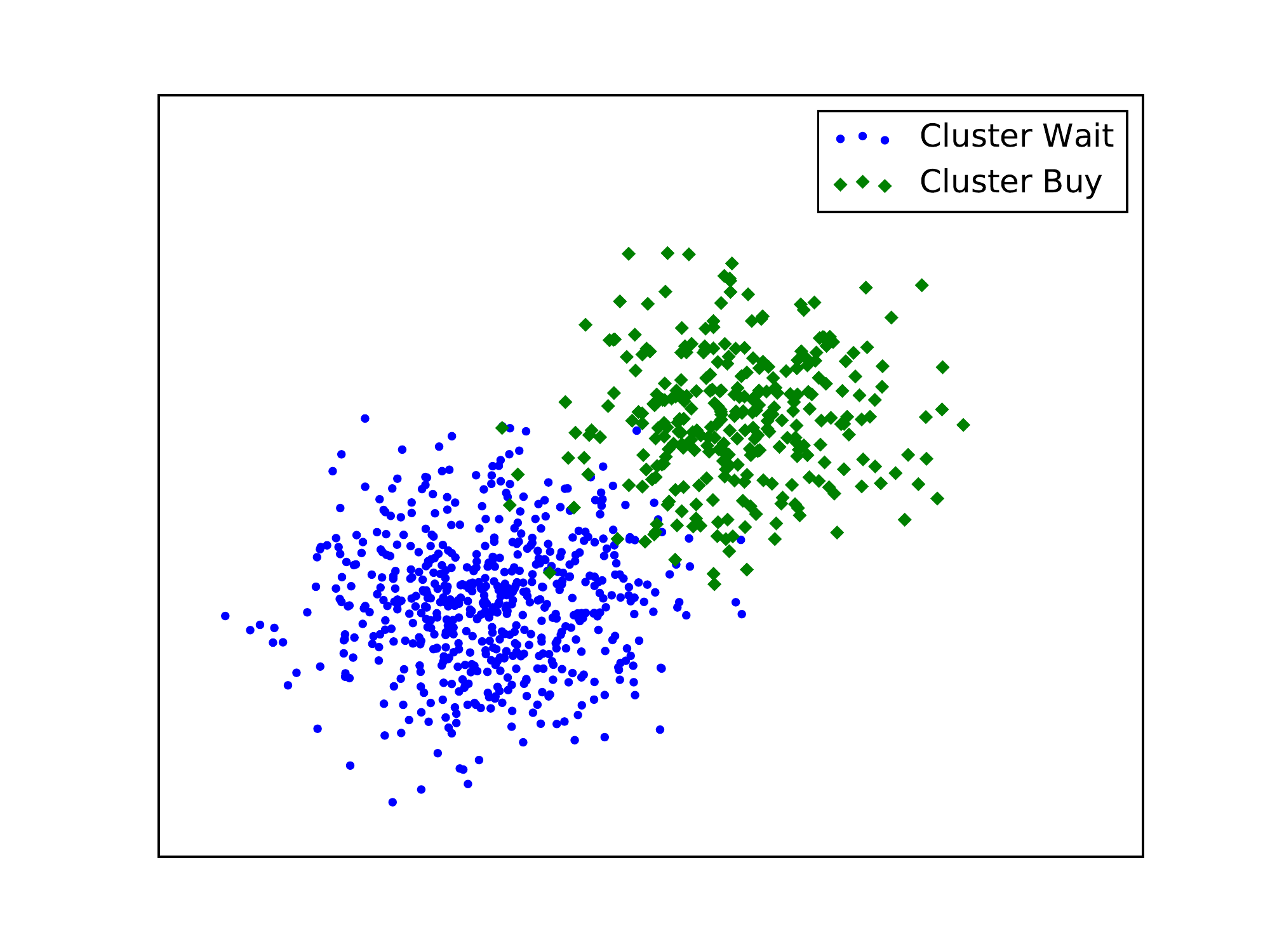}
\caption{Outlier Removal. }
\label{outlierRemoval}
\end{figure}

We addressed the problem of outlier removal by making use of unsupervised learning methods \cite{chawla2013k} \cite{liu2014unsupervised}, in particular through the implementation of K-Means and EM algorithm. Our approach was based on the fact that each characteristic class (either class 1 or 2) should in theory be restricted to a certain volume of the input space. Bearing this fact in mind, each of these two classes can be thought as a cluster. Consequently, we consider a sample to be an outlier if it does not belong to its labeled class cluster as illustrated by Fig. \ref{outlierRemoval}.

Having said that, we applied K-Means and EM to a training dataset. In the case of K-Means, 5896 samples were tagged as outliers. In the case of EM algorithm, 7318 samples were tagged as outliers. The choice of the initial conditions was crucial in terms of the algorithm’s convergence and in terms of obtaining a reliable result.

Regarding the initial conditions, for our propose of use, we computed the initial center to be such that: 

$\bf{\mu_k}^{initial} = \frac{1}{N_k} \sum_{\bf{x}_n\in C_k} \bf{x}_n $, where $N_k$ is the number of points belongs to Class k.

Fig \ref{outlierRemoval} shows how the outlier can be found due to clustering methods. The cluster buy entries evolve in the cluster wait should be considered to be outlier, and vice-versa.

We tested our classification methods on both the datas with and without outlier removal. Then we chose the best result to show.

\begin{table*}[h]
  \centering
\begin{tabular}{| c | c | c | c | c | c | c | c| c| c|} 
\hline Performance($\%$)& \multicolumn{9}{|c|}{Model}  \\
\cline{1-10} Routes $\backslash$ Method & Optimal  & \begin{tabular}[c]{@{}l@{}}Random\\ Purch.\end{tabular}  &  \begin{tabular}[c]{@{}l@{}}Logistic\\ Regression\end{tabular} & NN & \begin{tabular}[c]{@{}l@{}}Decision\\ Tree\end{tabular} & KNN & AdaBoost & \begin{tabular}[c]{@{}l@{}}Random\\ Forest\end{tabular} & \begin{tabular}[c]{@{}l@{}}Uniform\\ Blending\end{tabular}\\ 
\hline 
\hline 
R1 & 100.0 & 0.00 & 36.13 & 36.13 & 54.67 & 17.58 & 75.27 & 29.95 & 48.49\\ 
\hline 
R2 & 100.0 & 0.00 & 39.97 & 39.98 & 11.02 & 64.62 & 60.34 & 86.60 & 30.32\\ 
\hline 
R3 & 100.0 & 0.00 & 18.60 &18.60 & 70.51 & 83.48 & 71.69 & 52.81 & 45.73\\ 
\hline 
R4 & 100.0 & 0.00 & -22.16 & -22.15 & 57.76 & 38.35 & 32.64 & 0.67 & 46.34\\ 
\hline 
R5 & 100.0 & 0.00 & 53.25 & 53.26 & -26.42 & 34.44 & 50.76 & -34.16 & 65.15\\ 
\hline 
R6 & 100.0 & 0.00 & 57.30 & 57.30 & 80.35 & 87.69 & 85.33 & 19.83 & 71.71\\ 
\hline 
R7 & 100.0 & 0.00 & 31.03 & 31.03 & -28.2 & 55.28 & 65.02 & 33.61 & 49.12\\ 
\hline 
R8 & 100.0 & 0.00 & 64.63 & 64.63 & 23.63 & 14.99 & 49.76 & 9.57 & 57.8\\ 
\hline 
\hline
Mean Perf. & 100.0 & 0.00 & 34.84 & 34.84 & 30.42 & 49.55 & \bf{61.35} & 24.86 & 51.83\\ 
\hline
\hline
Variance & 100.0 & 0.00 & 660.74 & 660.74 & 1592.45 & 679.48 & 151.25 & 1148.16 & \bf{144.56}\\ 
\hline
\end{tabular}
 \caption{\bf{Classification Normalized Performance Comparison of 8 routes.}}
 \label{classificationResults}
\end{table*}

\subsubsection{\bf{Classification Performance Results}}
Table \ref{classificationResults} shows the results of classification methods. As we see, AdaBoost-DecisionTree, KNN, and Uniform Blending get positive performance for all the 8 routes and have smaller variance over these routes compared to other classification algorithms. The AdaBoost-DecisionTree method gets the best performance and a relative low variance over 8 routes. And as expected, the uniform blending method has the lowest variance just like the theory of uniform blending describes.

\subsection{Benchmark - \bf{Q Learning}}
We also implemented the method introduced in \cite{etzioni2003buy} to compare. Table \ref{qlearningResult} shows the result of Q-Learning. As we see, the Q-Learning method described in \cite{etzioni2003buy} has an acceptable performance and the variance is not large as well. The performance of it is very close to AdaBoost-DecisionTree Classification and Uniform blending Classification algorithms.

\begin{table}[h]
  \centering
\begin{tabular}{| c | c | c | c | c | c | c |} 
\hline Performance($\%$)& \multicolumn{3}{|c|}{Model}  \\
\cline{1-4}  & Optimal  & Random Purch. & Q Learning \\ 
\hline 
\hline 
R1 & 100.0 & 0.00 & 68.76 \\ 
\hline 
R2 & 100.0 & 0.00 & 61.81\\ 
\hline 
R3 & 100.0 & 0.00 & 51.13 \\ 
\hline 
R4 & 100.0 & 0.00 & 54.01 \\ 
\hline 
R5 & 100.0 & 0.00 &  61.27\\ 
\hline 
R6 & 100.0 & 0.00 &  72.08\\ 
\hline 
R7 & 100.0 & 0.00 &  65.29\\ 
\hline 
R8 & 100.0 & 0.00 &  6.50\\ 
\hline 
\hline
Mean Perf. & 100.0 & 0.00 & 55.11\\ 
\hline
\hline
Variance & 0.00 & 0.00 & 380.06\\ 
\hline
\end{tabular}
 \caption{\bf{Q Learning Performance.}}
 \label{qlearningResult}
\end{table}

\subsection{\bf{Generalized Problem Performance Result}}
Table \ref{generalResult} shows the result of generalized problem. As we can see, the uniform blending does not get any improvement. But the HMM Sequence Classification algorithm makes 9 routes get improvement, 3 routes have negative performance. Although the average performance is 31.71$\%$, which is lower than that of the specific problem, it makes sense that we did not use any historical data of these routes to predict (actually, there are two routes already appear in the specific problem, which are R3=R11 and R6=R16). In specific problem, when using the AdaBoost-DecisionTree Classification, the performances for these two routes (i.e. R3 and R6) are 71.69$\%$ and 85.33$\%$ respectively. However, in generalized problem, using same classification method, the performances are 63.11$\%$ and 0.54$\%$, which has poorer performance than the specific problem. This is tolerable because we only used the formula trained in specific problem to predict.

\begin{table}[h]
  \centering
\begin{tabular}{| c | c | c | c | c | c | c | c|} 
\hline Perf. ($\%$)& \multicolumn{4}{|c|}{Model}  \\
\cline{1-5}  & Optimal  & \begin{tabular}[c]{@{}l@{}}Random\\ Purch.\end{tabular} & Uniform & HMM \\ 
\hline 
\hline 
R9 & 100.0 & 0.00 & 80.52 & 87.09\\ 
\hline 
R10 & 100.0 & 0.00 & -75.1 & -48.6\\ 
\hline 
R11 & 100.0 & 0.00 & 51.06 & 63.11\\ 
\hline 
R12 & 100.0 & 0.00 & -16.57 & 17.32\\ 
\hline 
R13 & 100.0 & 0.00 &  9.47 & 53.22\\ 
\hline 
R14 & 100.0 & 0.00 &  21.14 & 45.83\\ 
\hline 
R15 & 100.0 & 0.00 & 24.95 & 62.33\\ 
\hline 
R16 & 100.0 & 0.00 &  -56.99 & 0.54\\ 
\hline 
R17 & 100.0 & 0.00 &  -105.07 & -84.45 \\ 
\hline 
R18 & 100.0 & 0.00 &  -0.18 & 38.16\\ 
\hline 
R19 & 100.0 & 0.00 &  -48.61 & -15.97\\ 
\hline 
R20 & 100.0 & 0.00 &   -3.3 & 35.12\\ 
\hline 
\hline
Mean Perf. & 100.0 & 0.00 & -14.84 & 31.71\\ 
\hline
\hline
Variance & 0.00 & 0.00 & 2637.84 & 2313.25\\ 
\hline
\end{tabular}
 \caption{\bf{Generalized Model Performance.}}
 \label{generalResult}
\end{table}

\section{\bf{Conclusion}}
In this article, we used the airplane ticket data over a 103 day period for 8 routes to perform out models. Removing outlier through K-Means Algorithm and EM Algorithm implied that our training algorithms were not influenced by non-representative class members; tackling the fact that our dataset was imbalanced, through Random Over Sampling, meant our algorithms were not biased towards the majority classes. For the classification and regression methods, the best values for hyperparameters were found through 5-fold grid search. 

As shown by the results, for the \textbf{specific problem} and from the aspect of performance, AdaBoost-Decision Tree Classification is suggested to be the best model, which has \textbf{61.35}$\%$ better performance over random purchase strategy and has relatively small performance variance for the 8 different routes. From the aspect of performance variance for different routes, Uniform Blending Classification is chosen as the best model with relatively high performance. On the other hand, the Q-Learning method got a relatively high performance as well. Compare the results of regression methods and classification methods, we could find that the cross validation error or precision in classification has far smaller variance than that of regression. We then considered that the classification model construction is more suitable in this problem. 


For the \textbf{generalized problem} (i.e. predict without the historical data of routes that we want to predict), we did not test many models. However, the HMM Sequence Classification based AdaBoost-Decision Tree Classification model got a good performance over 12 new routes, which has \textbf{31.71}$\%$ better performance than the random purchase strategy. 


For the generalized problem, we only used two methods to predict for generalized routes. In the future, we may find more algorithms to see how can we extend the ticket prediction to generalized routes. Because this model may have many benefits, such as reducing computation time.

\bibliographystyle{named}
\bibliography{bib}

\end{document}